\documentclass{article}

\usepackage[preprint,nonatbib]{nips_2018}

\usepackage[utf8]{inputenc} 
\usepackage[T1]{fontenc}    
\usepackage{hyperref}       
\usepackage{url}            
\usepackage{booktabs}       
\usepackage{amsfonts}       
\usepackage{nicefrac}       
\usepackage{microtype}      
\usepackage{graphicx}
\usepackage{epsfig}
\usepackage{amsmath,amsthm,amssymb}
\usepackage{algorithm}
\usepackage[noend]{algpseudocode}

\usepackage{bm}

\newcommand{\excise}[1]{}
\newcommand{\R}[1]{\mathbb{R}^{#1}}

\newcommand{\myvector}[1]{\bm{#1}}
\newcommand{\myvec}[1]{\myvector{#1}}

\newcommand{\xl}[1]{\bm{x}^{#1}}

\newcommand{\cf}{c}

\title{MPDCompress - Matrix Permutation Decomposition Algorithm for Deep Neural Network Compression}

\author{Lazar Supic \\
Department of Computer Science\\
University of California, Berkeley\\
Berkeley, CA 94720, USA \\
\texttt{lazar@berkeley.edu}\\
\And
Rawan Naous\\
Department of Computer Science\\
University of California, Berkeley\\
Berkeley, CA 94720, USA \\
\texttt{ rawansn@berkeley.edu}
\And
Ranko Sredojevic\\
Department of Computer Science\\
University of California, Berkeley\\
Berkeley, CA 94720, USA \\
\texttt{rrs@berkeley.edu}
\And
Aleksandra Faust\\
Google Brain\\
Mountain View, CA, 94025, USA\\
E-mail: faust@google.com \\
\And
Vladimir Stojanovic\\
Department of Computer Science\\
University of California, Berkeley\\
Berkeley, CA 94720, USA \\
\texttt{vlada@berkeley.edu}
}

\begin{document}

\maketitle

\begin{abstract}
Deep neural networks (DNNs) have become the state-of-the-art technique for machine learning tasks in various applications. However, due to their size and the computational complexity, large DNNs are not readily deployable on edge devices in real-time. To manage complexity and accelerate computation, network compression techniques based on pruning and quantization have been proposed and shown to be effective in reducing network size. However, such network compression can result in irregular matrix structures that are mismatched with modern hardware-accelerated platforms, such as graphics processing units (GPUs) designed to perform the DNN matrix multiplications in a structured (block-based) way. We propose MPDCompress, a DNN compression algorithm based on matrix permutation decomposition via random mask generation. In-training application of the masks molds the synaptic weight connection matrix to a sub-graph separation format. Aided by the  random permutations, a hardware-desirable block matrix is generated, allowing for a more efficient implementation and compression of the network. To show versatility, we empirically verify MPDCompress on several network models, compression rates, and image datasets. On the LeNet 300-100 model (MNIST dataset), Deep MNIST, and CIFAR10, we achieve 10$\times$ network compression with less than 1$\%$ accuracy loss compared to non-compressed accuracy performance. On AlexNet for the full ImageNet ILSVRC-2012 dataset, we achieve 8$\times$ network compression with less than 1$\%$ accuracy loss, with top-5 and top-1 accuracies of 79.6$\%$ and 56.4$\%$, respectively. Finally, we observe that the algorithm can offer inference speedups across various hardware platforms, with 4$\times$ faster operation achieved on several mobile GPUs.
\end{abstract}



\maketitle

Deep neural networks (DNNs) have become the leading technique used in machine learning tasks across multiple application domains in computer vision \cite{StructuredPrunning2,qi2016pointnet,toshev2017deeppose}, ranging from object perception in autonomous vehicles \cite{bechtel2017picar,smolyanskiy2017autoMAV,smolyanskiy2018stereo} to medical imaging and diagnosis \cite{esteva2017dermatologistlevel, SqueezeNet, ImageNet, LeCunNature,MedImaging,VeryDeepCNN}. However, due to their size and complexity, large DNNs can be quite computationally intensive, which can pose a significant challenge for real-time applications on edge devices \cite{Energy_aware_pruning,StructuredPrunning2,Dropout}. Specifically, state-of-the art DNNs can contain hundreds of layers represented by matrices containing millions of elements, such that both the memory requirements and the computational complexity of matrix multiplication needed for network training and high-accuracy performance rapidly increases with DNN size \cite{SqueezeNet,ImageNet,Convolutions,DeepResidual}. This challenge is particularly difficult for DNNs containing fully-connected (FC) layers characterized by large matrix sizes \cite{SqueezeNet,ImageNet}. DNNs with large FC layers can be found in over 60$\%$ of production-side applications running in a cloud services platform, highlighting the practical importance of this challenge \cite{TPU}. To create a more perfect union between network compression and the underlying compute platform, we propose a matrix permutation decomposition algorithm, termed MPDCompress, that both compresses DNNs and molds them to the underlying hardware-accelerated platform. 
We experimentally evaluate our MPDCompress algorithm on several DNN models, including LeNet 300-100 (MNIST dataset), Deep MNIST, CIFAR10, and AlexNet (ImageNet ILSVRC-2012 dataset) and achieve less than 1$\%$ accuracy loss compared to non-compressed accuracy performance. 
In the next section, we review related work on network compression and discuss the key challenges for irregular matrix structures that motivated this work. The following sections explain matrix permutation decomposition via random binary mask generation and application, introduce our algorithm architecture for both training and inference modes, and describe our experimental framework. We also present our experimental results on several different DNNs and datasets to confirm the versatility of the approach. Finally, we observe that MPDCompress can offer inference speedups across various hardware platforms, with 4$\times$ faster operation on several mobile GPUs~\cite{structuredDeep}.

\section{Related Work}
To manage growing DNN complexity, algorithmic methods for network compression (e.g. pruning, quantization) \cite{Han1,HanDSD,HanESE,Noiseout,DropNeuron, TernaryQuant,StructuredPrunning1,CustomizingDNN,HubaraQuantizedNN,SEPNet,RandomizedHashing,pmlr-v37-chenc15,Wen_NIPS2016,kang2017shakeout,Ullrich2017ICLR,kim2016mobileapp} have been proposed and demonstrated. Quantization techniques mainly focus on reducing the number of bits used to represent the matrix coefficients, while pruning techniques rely on the intrinsic sparsity of the matrix coefficients within a network layer.  Pruning-based network compression thus allows for fully removing nodes and connections with small coefficients since they do not significantly contribute to the output value of the computation. From the algorithmic perspective, the key steps in pruning and quantization based network compression approaches include first training a given DNN on a given data set, next performing compression via pruning and quantization that are built around “important connections” (i.e. large matrix coefficients that arise during the first training step), and then re-training. These approaches have shown to be very effective in compressing network size. For example, in \cite{Han1}, weights that are below a certain threshold are directly removed, achieving around 9$\times$ to 13$\times$ compression rates. In \cite{StructuredPrunning1}, the important connections are determined through the use of sequential Monte Carlo method or particle filters.  Alternately, \cite{CustomizingDNN} uses SIMD-aware weight pruning along with node pruning to reach around 2$\times$ to 5$\times$  compression while also addressing the hardware platform, with different approaches applied for micro-controllers, CPUs and GPUs respectively. 

A remaining challenge is that in terms of network structure, compression via pruning and quantization can result in irregular matrix structures that can be mismatched with modern compute platforms. Specifically, fully connected layers can be made very sparse via network compression, but the sparse values tend to be scattered irregularly, such that a large matrix would still have to be stored in memory. Moreover, the processor would need to be alerted with extra flags and pointers to the locations of non-zero coefficients within the large matrix, which can notably reduce the effective level of compression and speedup achieved for a particular pruning technique \cite{CustomizingDNN}. Consequently, an algorithm that can "pack" the matrix content into the most efficient form, both in terms of memory storage and compute, is desirable. The optimal packing form is governed by the  underlying computational hardware, which in this case seeks to divide matrix computations into blocks that it processes in parallel (e.g. NVIDIA GPU). Dense blocks of non-zero values also happen to be the best packing form for efficient GPU memory usage \cite{efficient_memory}.

In this work, we present MPDCompress, a matrix permutation decomposition algorithm that invokes random permutations of rows and columns of matrices representing FC layers to create compressed DNNs that can be tailored (i.e. molded) to the underlying hardware platform structure. We show that the proposed algorithm “forces” the final computational representation of the neural network structure to be in the block diagonal form, and as such, it becomes particularly favorable to parallelized GPU-based compute platforms. The key enabler introduced by our algorithm is the creation of independent blocks for matrix computations, such that the matrix multiplication and accumulation required for each block of weights and activation parameters has no dependence on any other blocks. This split allows for a parallel implementation of the required operations and can thus speed up the overall inference time. We also note that whereas previous pruning techniques available in the literature \cite{Han1} explicitly rely on the existence and preservations of important connections that emerge after the training phase, our MPDCompress algorithm requires no fixed important connections. The random permutation masks allow for synaptic connections to be set randomly and pruned accordingly. An example of this will be demonstrated in the results section, where the accuracy of the compressed network is preserved using 100 different masks within the context of the MPDCompress algorithm.
\section{MPDCompress Algorithm}

\begin{figure}[h]
\begin{center}
   \includegraphics[width=1\linewidth]{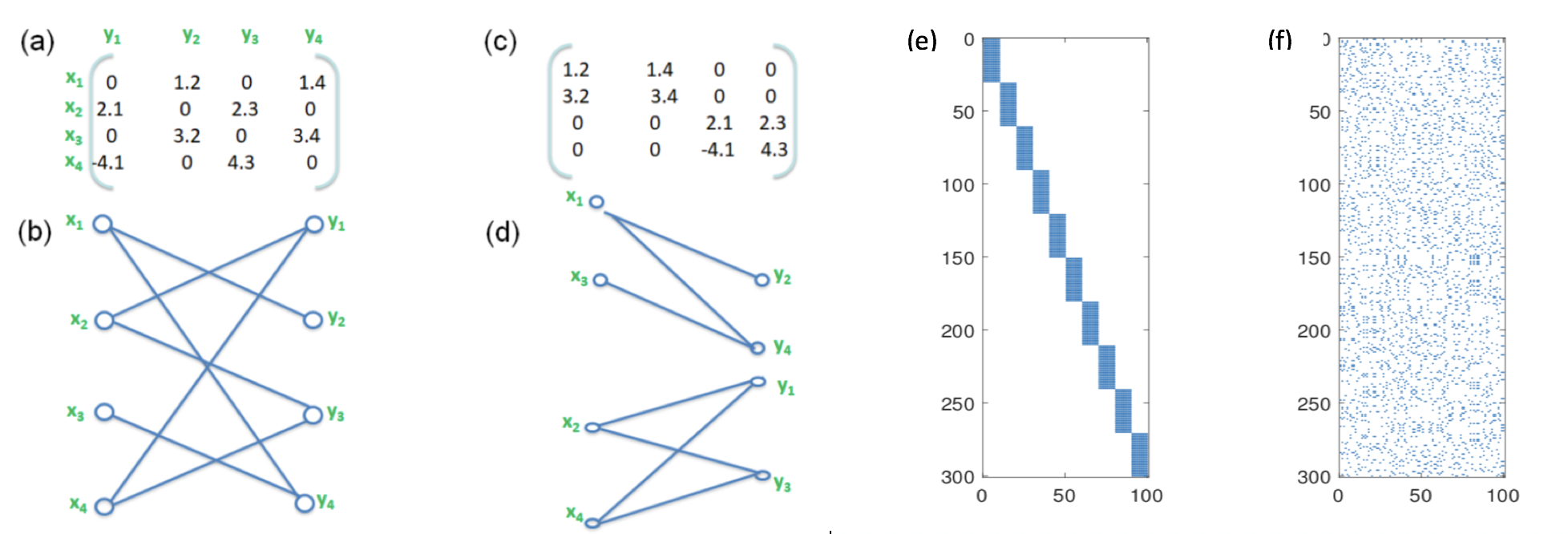}
\end{center}
   \caption{(a) Example of a 4$\times$4 sparse matrix with irregular structure; (b) graph representation of the 4$\times$4 matrix of Fig. \ref{fig1}(a); (c) block-diagonalized version of the 4$\times$4 matrix of Fig. \ref{fig1}(a), achieved by row and column permutations; (d) graph representation of the block-diagonalized matrix of Fig. \ref{fig1}(c). (e) 300$\times$100 block-diagonal matrix, $\textbf{B}_1$; (f, right) 300$\times$100 binary mask, $\textbf{M}_1$ created by randomly permuting the rows and columns of $\textbf{B}_1$.}
\label{fig1}
\end{figure}

The key motivation for the proposed algorithm is that matrices representing fully-connected layers can be made very sparse, but these sparse values tend to be distributed irregularly, such that it becomes necessary to store the entire large sparse matrix in memory despite relatively few non-zero values in it. Consequently, it is desirable to have an algorithm that can pack the non-zero values into a more compact form that is efficient both in terms of memory storage and compute. This is particularly true for fully connected (FC) layers, which, due to the nature of their connections, cannot significantly benefit from data reuse or numerical optimization methods \cite{TPU}. For a GPU-based platform, for example, inducing a block structure onto the matrix representing the FC layers both streamlines matrix computations and increases memory utilization efficiency. 

The key question then becomes how to induce a block structure onto an irregularly-structured sparse matrix. To first show that this is feasible, we invoke the graph representation for an example 4$\times$4 sparse irregular matrix shown in Fig. \ref{fig1}(a). In Fig. \ref{fig1}(b), a graph connection is made between row and column elements $x_i$ and $y_i$ if there is a non-zero value in the ($x_i, y_i$) location of the 4$\times$4 matrix. As shown in Fig. \ref{fig1}(d), independent sub-graphs emerge in the graph representation of the original 4$\times$4 matrix, wherein each of these sub-graphs defines a 2$\times$2 block in an equivalent 4$\times$4 block-diagonal matrix Fig. \ref{fig1}(c).  The subgraphs also define the matrix row and column permutations needed to put the original 4$\times$4 matrix into the block form. Specifically, by applying the row and column permutations $[{\textbf{x}_1, \textbf{x}_3, \textbf{y}_2, \textbf{y}_4}];
[{\textbf{x}_2, \textbf{x}_4, \textbf{y}_1, \textbf{y}_3}],$
the original irregular sparse 4$\times$4 matrix of Fig. \ref{fig1}(a) is decomposed to the block-diagonal matrix of Fig. \ref{fig1}(c). We thus conclude that if sub-graph separation may be observed in the graph representation of the original irregular sparse matrix, the matrix can be decomposed into a block form through permutations of its rows and columns. 

The next question therefore is how to ensure that an arbitrary irregular sparse matrix will have the needed sub-graph separation. In the proposed algorithm, we achieve this by applying a binary mask to such a matrix, wherein the mask is created by randomly permuting the rows and columns of a block-diagonal matrix.
As an example, Fig. \ref{fig1}(e) shows a 300$\times$100 block-diagonal matrix, $\textbf{B}_1$, containing 3000 non-zero elements (10$\%$ matrix sparsity), while Fig. \ref{fig1}(f) shows the 300$\times$100 binary mask, $\textbf{M}_1$, created by randomly permuting the rows and columns of $\textbf{B}_1$. Since the block-diagonal matrix, $\textbf{B}_1$, used to create the binary mask $\textbf{M}_1$, has sub-graph separation by definition, an output matrix achieved by applying $\textbf{M}_1$ onto an arbitrary irregular sparse matrix will also have the necessary sub-graph separation. By then performing the inverse of the random permutation used to generate the mask, we induce the desired block-diagonal structure onto the output matrix. 
\begin{figure*}[h]
\begin{center}
\includegraphics[width=1\linewidth]{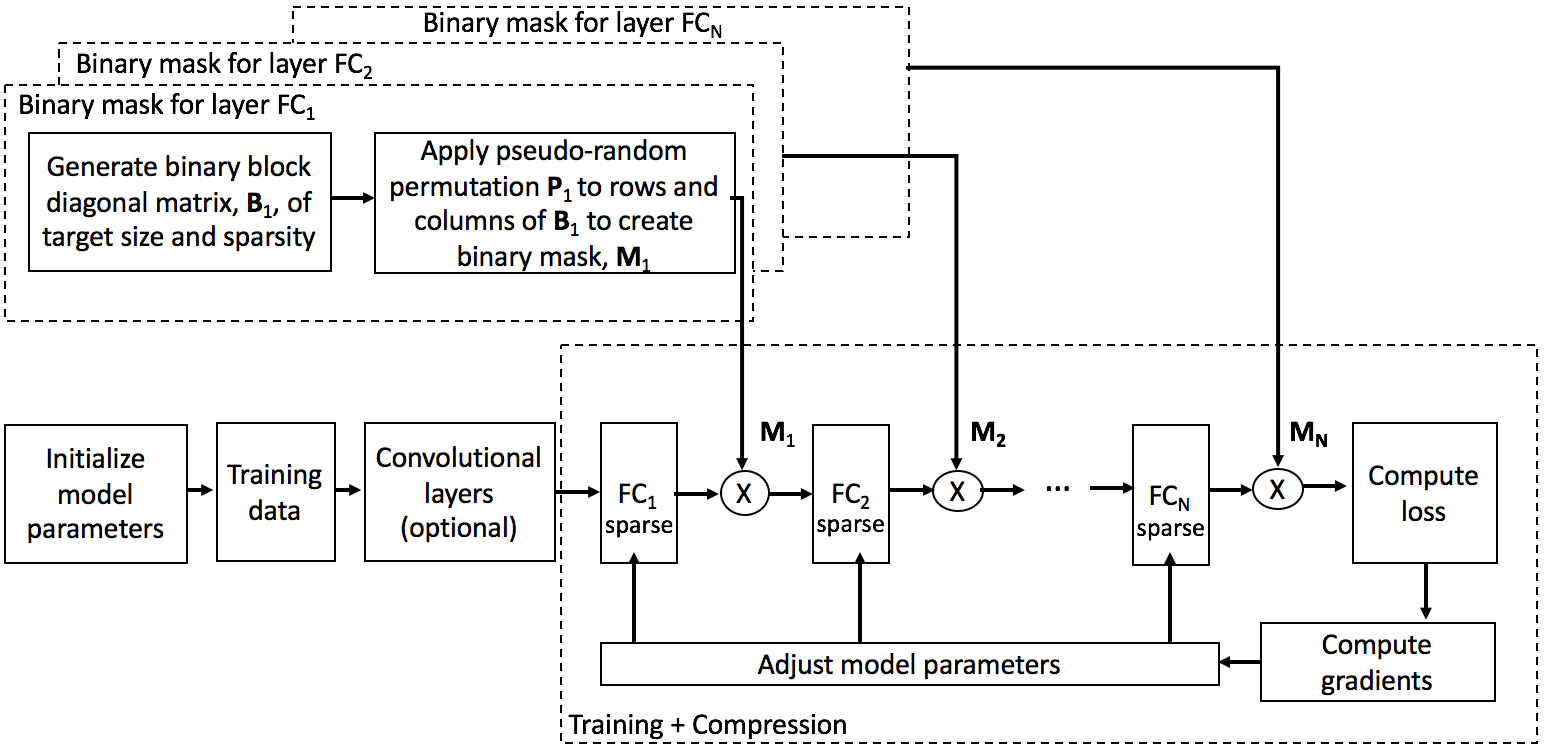}
\end{center}
   \caption{Matrix permutation decomposition algorithm architecture in training + compression mode}
\label{fig3}
\end{figure*}

\begin{figure*}[h]
\begin{center}
\includegraphics[width=1\linewidth]{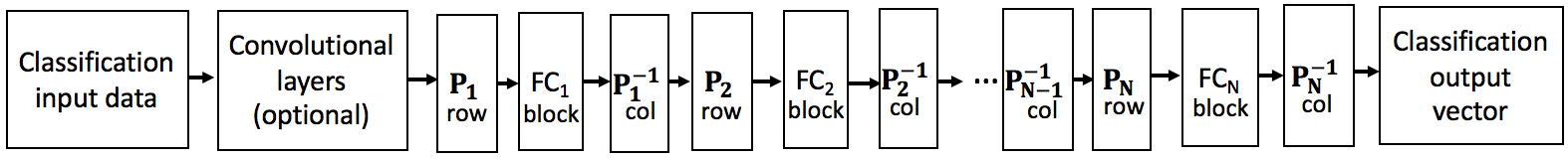}
\end{center}
   \caption{Matrix permutation decomposition algorithm architecture in inference mode.}
\label{fig4}
\end{figure*}


Fig. \ref{fig3} builds on the random matrix permutation concepts introduced in Fig. \ref{fig1}, and illustrates the proposed network compression algorithm as used in training + compression  mode. As shown in Fig. \ref{fig3}, to create a binary mask for each FC layer of the DNN, we first start with a block diagonal matrix, $\textbf{B}_i$, $i$ = 1, ..., $N$, whose size is determined by the size of the fully-connected layer to which it will be applied. The level of sparsity for $\textbf{B}_i$ is controlled as a hyper parameter. As mentioned above, this matrix will have binary non-zero blocks along the main diagonal, with other values set to zero. Next, we create the binary mask, $\textbf{M}_i$ for fully-connected layer FC$_{i}$ by applying a random permutation, $\textbf{P}_i$ to the rows and columns ($\textbf{P}_{i,row}$ and $\textbf{P}_{i,col}$, respectively) of $\textbf{B}_i$. We note that the mask $\textbf{M}_i$ is only applied to the weight matrix, as the biases are added constants and could be integrated within the computations. We also note that it is sufficient to generate a single random mask per FC layer, and have confirmed via experiment that different random mask instantiations do not significantly affect classification accuracy; these results will be presented in the following section. As further shown in Fig. \ref{fig3}, once a binary mask is created for each of $N$ FC layers in the network, the binary masks are applied to the corresponding FC layers on each training iteration by multiplying the matrix coefficients of the given FC layer with the mask. This process is iterated over the training dataset by computing the resulting loss and gradients and adjusting FC layer coefficients until the desired accuracy is achieved. It is note worthy that binary masks are applied only on the updated weights after the gradient descent calculation. We note that each iteration comprises a gradient descent iteration that will force the neural network coefficients in the FC layers to align with the binary mask structures. In this way, we seek to reconstruct a neural network model that has the hardware-favorable block diagonal structure during the inference stage. 
%
%

Fig. \ref{fig4} illustrates the operation of our algorithm in inference mode. After the classification data is filtered through (optional) convolutional layers, it becomes input to the first FC layer, FC$_{1}$, which is block-diagonalized through the application of the random permutation $\textbf{P}_1$ onto it; we recall that $\textbf{P}_1$ is the same permutation used to generate the binary mask $\textbf{M}_1$ in Fig. \ref{fig3}. 
such that the matrix multiplications needed to pass the classification data through this layer can be done in a highly memory- and compute- efficient way. 
This process is repeated for all $N$ FC layers in the network, culminating with the final classification output vector, as shown in Fig. \ref{fig4}, from which inference accuracy can be obtained. Also, the row and column components of the permutations for consecutive layers, e.g. $\textbf{P}^{-1}_{1,col}$ and $\textbf{P}_{2,row}$, could be the inverses of each other, thus forming the identity matrix and eliminating the need for internal permutations.

Pseudo code for MPDCompress is shown in the Algorithm 1 listing. The listing details the generation of the masks in a binary format that is easily mapped into the training phase with a mere multiplication step to the weight matrix. The application of the decomposition algorithm is performed per batch with no limitation on either the batch size or the training procedure.

\makeatletter
\def\BState{\State\hskip-\ALG@thistlm}
\makeatother

\begin{algorithm}
\caption{MPDCompress}\label{euclid}
\begin{algorithmic}[1]
\Procedure{Creating Masks}{}
\State $\textit{nFC} \gets \text{number of }\textit{FC layers in DNN}$
\State $i \gets \textit{1}$
\For{\texttt{i <= nFC}}
        \State \text{determine dimensions of ith FC layer}
        \State \text{create block diagonal mask for i$^{th}$ layer} $\textbf{B}_i$
        \State \text{create permutation $\textbf{P}_i$ for i$^{th}$ layer} 
        \State \text{create binary mask $\textbf{M}_i$ for i$^{th}$ layer}, applying $\textbf{P}_i$ to $\textbf{B}_i$

        \State $i \gets i+1$.
\EndFor
\EndProcedure
\Procedure{DNN Training with MPD Algorithm}{}
\State \text{initialize weights and biases}
\State \text{import the masks to the graph} $\textbf{M}_i$
\BState \emph{per batch}:
	\State \text{multiply binary mask $\textbf{M}_i$ with the weight matrix  $\textbf{W}_i$} 
        \State \text{supply input images and activations for the layers}
        \State \text{calculate the gradient and update the weights}

\EndProcedure
\end{algorithmic}
\end{algorithm}

Mathematically, the application of the block-permuted matrices to a fully connected neural network per the MPDCompress algorithm may be expressed as follows: let the $n$-layer FC DNN be defined as \[X = [\xl{1} \cdots \xl{n}] \]  where  \[\xl{i} \in \R{d_i}, \] 
\[\xl{i+1} = f_i(\textbf{W}_i\xl{i} + \myvec{b}^{i})\]
a\\$i=1,..,n,$ $f_i(\xl{}) = \max(0, \xl{})$ is a RELU activation function at layer $i$, $\textbf{W}_{i} \in \R{d_{i+1} \times d_{i}}$, $d_{i}$ is the number of neurons in the $i^{th}$ DNN layer, and $d_{i+1} \times d_{i}$ are dimensions of the weight matrix $\textbf{W}_{i}$.
We define a random mask with compression factor, $c_i$ for the $i^{th}$ layer as
\[\textbf{M}^{d_{i+1}, d_{i}}_{\cf_i} = \textbf{P}_{i,row}(d_{i+1}) \textbf{B}^{d_{i+1}, d_{i}}_{\cf_i} \textbf{P}_{i,col}(d_{i}).\]
The mask is created by randomly permuting the rows and columns of block diagonal identity matrix compatible with the $i^{th}$ layer, $\textbf{B}^{d_{i+1}, d_{i}}_{\cf_i}$. 

We create a sparse weight matrix at layer $i$ by applying the mask to the weight matrix,
\begin{equation}
\overline{\textbf{W}_{i}} =\textbf{M}^{d_{i+1}, d_{i}}_{\cf_i} \circ \textbf{W}_{i}.
\end{equation}
Since the $\textbf{M}^{d_{i+1}, d_{i}}_{\cf_i}$ is a permutation of a block-diagonal matrix $$\textbf{M}^{d_{i+1}, d_{i}}_{\cf_i} \sim \textbf{B}^{d_{i+1}, d_{i}}_{\cf_i},$$ the sparse layer matrix $\overline{\textbf{W}}_{i}$ is also a permutation of the block diagonal matrix $$\overline{\textbf{W}}_{i} \sim \textbf{P}^{\intercal}_{i,row}(d_{i+1}) \textbf{W}_{i} \textbf{P}^{\intercal}_{i,col}(d_{i}) \circ \textbf{B}^{d_{i+1}, d_{i}}_{\cf_i}$$ and has the necessary sub-graph separation. 

Thus, to train a fully-connected layer we use:
\[ \xl{i+1} = f_i(\overline{\textbf{W}}_{i}\xl{i} + \myvec{b}^{i})\]

At inference time, we induce the desired block-diagonal structure onto the output matrix by performing the inverse permutation used to generate the mask, 
\begin{equation}
\textbf{W}^*_{i} = \textbf{P}^{\intercal}_{i,row}(d_{i+1}) \overline{\textbf{W}}_{i} \textbf{P}^{\intercal}_{i,col}(d_{i}).
\end{equation}
We observe that $\textbf{W}^*_{i}$ is sparse, and block diagonal. We thus need to transform the inputs and outputs of the layers to use $\textbf{W}^*_{i}$ instead of $\overline{\textbf{W}}_{i}.$ Next, we derive the transformation.
\[ \xl{i+1} = \textbf{P}_{i,row}(d_{i+1}) f_i(\textbf{W}^*_{i} \xl{i} + \myvec{b}^{i})\textbf{P}_{i,col}(d_{i})\]
\begin{align*}
\xl{i+1} &= f_i(\overline{\textbf{W}}_{i} \textbf{P}^{\intercal}_{i,col}
(d_{i})  \textbf{P}_{i,col}(d_{i}) \xl{i} + \myvec{b}^{i}) = f_i(\textbf{P}_{i,row}(d_{i+1}) \textbf{P}^{\intercal}_{i,row}(d_{i+1}) \overline{\textbf{W}}_{i} \textbf{P}^{\intercal}_{i,col}(d_{i})\textbf{P}_{i,col}(d_{i})  \xl{i+1}
\myvec{b}^{i})\\
&= f_i(\textbf{P}_{i,row}(d_{i+1}) \textbf{W}^*_{i}  \textbf{P}_{i,col}
(d_{i}) \xl{i} + \myvec{b}^{i})\\
&= f_i(\textbf{P}_{i,row}(d_{i+1}) \textbf{W}^*_{i}  \textbf{P}_{i,col}
(d_{i}) \xl{i}+
\textbf{P}_{i,row}(d_{i+1}) 
\textbf{P}^{\intercal}_{i,row}(d_{i+1}) 
\myvec{b}^{i})\\
&= \textbf{P}_{i,row}(d_{i+1}) \max(0, \textbf{W}^*_{i}  \textbf{P}_{i,col}(d_{i}) \xl{i} + \textbf{P}^{\intercal}_{i,row}(d_{i+1}) \myvec{b}^{i})\\
&= \textbf{P}_{i,row}(d_{i+1}) \max(0, \textbf{W}^*_{i} \xl{i}_{\textbf{P}_{i,col}}+\myvec{b}^{i}_{\textbf{P}^{\intercal}_{i,row}}),\\
\end{align*}
where $\xl{i}_{\textbf{P}_{i,col}} = \textbf{P}_{i,col}(d_{i}) \xl{i}$ and $\myvec{b}^{i}_{\textbf{P}^{\intercal}_{i,row}} = \textbf{P}^{\intercal}_{i,row}(d_{i+1}) \myvec{b}^{i}.$
\begin{align*}
\textbf{P}^{\intercal}_{i,row}(d_{i+1}) \xl{i+1} &= \max(0, \textbf{W}^*_{i}  \xl{i}_{\textbf{P}_{i,col}} + \myvec{b}^{i}_{\textbf{P}^{\intercal}_{i,row}})\\
\xl{i+1}_{\textbf{P}^{\intercal}_{i,row}} &= \max(0, \textbf{W}^*_{i}  \xl{i}_{\textbf{P}_{i,col}} + \myvec{b}^{i}_{\textbf{P}^{\intercal}_{i,row}}),
\end{align*}
and $\xl{i+1}_{\textbf{P}^{\intercal}_{i,row}} = \textbf{P}^{\intercal}_{i,row}(d_{i+1}) \xl{i+1}.$
Let $\textbf{P}_{i+1,row} = \textbf{P}_{i,row}(d_{i+1}) \in \R{d_{i+1} \times d_i}$ be the unpacking matrix on layer $i$ output $\xl{i+1},$ and $\textbf{P}_{i,col}= \textbf{P}_{i,col}(d_{i}) \in \R{d_{i} \times d_{i}},$ be an unpacking matrix on $i^{th}$ layer's input, $\xl{i}.$  This concludes the mathematical description of the algorithm.

\section{Experimental Results}
We experimentally evaluated our network compression algorithm on several different DNN models and datasets of various sizes and complexity levels. Specifically, the DNN models we used include the LeNet 300-100 (MNIST dataset), AlexNet (ImageNet ILSVRC-2012 dataset), Deep MNIST, and CIFAR10. Our experimental evaluations were performed using Tensorflow as the simulation platform into which we inserted our network compression algorithm first during training (Fig. \ref{fig3}) and then during inference (Fig. \ref{fig4}). We discuss the results obtained for LeNet 300-100 and AlexNet in more detail in this section, while the results obtained for Deep MNIST and CIFAR10 are summarized in Table 2.

\subsection{LeNet 300-100 with MNIST}
\begin{figure}[t]
\begin{center}
   \includegraphics[width=1\linewidth]{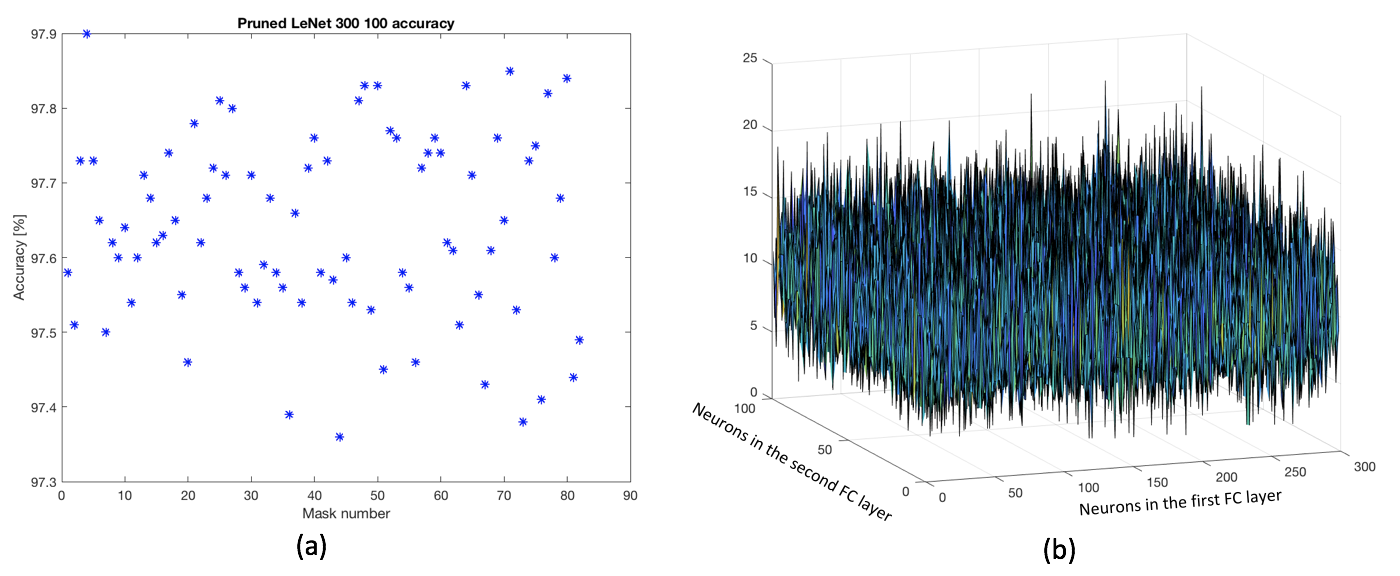}
\end{center}
   \caption{(a, left) LeNet300-100 network accuracy achieved for each of the 100 masks; (b, right) Sum of 100 different masks across the second fully connected layers of LeNet 300-100.}
\label{fig6}
\end{figure}

We first performed the experimental evaluation of the proposed network compression algorithm on the first and second fully connected layers of the LeNet 300-100 DNN for the MNIST dataset of hand written digits (28$\times$28 pixels). Out of the 60k images in the MNIST dataset, 10k were used for evaluation. We used a minibatch size of 50 images and fixed learning rate of 10$^{-3}$. The sparsity level was set to 10$\%$, with 784$\times$300 and 300$\times$100 random binary masks applied to the first and second layers of the DNN, respectively. Training accuracy of 98.14$\%$ was achieved, and inference accuracy was assessed for 100 different masks per layer generated via random permutations as described above.  As shown in Fig. \ref{fig6}(a), accuracy greater than 97.3$\%$ was achieved for each of the 100 masks,  which results in less than 1$\%$ accuracy loss compared to the non-compressed LeNet300-100 network accuracy of 98.16$\%$. To further assess the random nature of the masks, we summed the 100 different masks for both layers and plotted the 3D sum as shown in Fig. \ref{fig6}(b). As the masks are binary, the sum represents masks that share non-zero values in the same matrix location. The sum on average reached 10, confirming the high spread of non-zero mask values across the matrix. To highlight the role of random permutations in the mask generation process, accuracy was also evaluated by using non-permuted block-diagonal binary masks  with 10$\%$ sparsity. Using the non-permuted masks, we achieved 80.2$\%$ accuracy with 10$\%$ sparsity, versus above 97$\%$ accuracy for random mask permutations with 10$\%$ sparsity (Fig. \ref{fig6}(a)). By increasing sparsity to 20$\%$, we achieved 85.97$\%$ accuracy using non-permuted block diagonal masks. Consequently, we observe that random permutations in mask generation better preserve information flow from layer-to-layer compared to non-permuted mask application.


\subsection{AlexNet on ImageNet}
\begin{figure}
\begin{center}
   \includegraphics[width=1\linewidth]{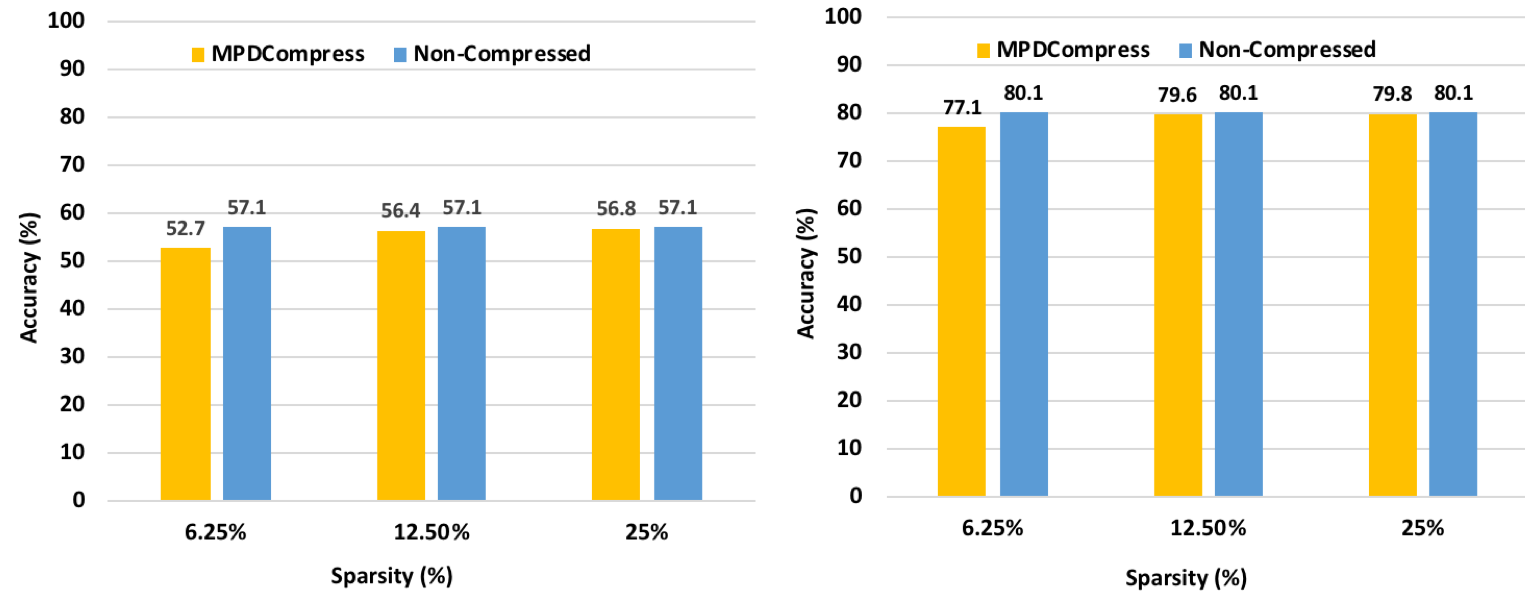}
\end{center}
   \caption{(a, left) Top1 accuracy experimental results for AlexNet on ImageNet achieved for different levels of network sparsity using MPDCompress; (b, right) Top5 accuracy experimental results for AlexNet on ImageNet achieved for different levels of network sparsity using MPDCompress. Comparison is provided with respect to non-compressed accuracy results.}
\label{Top1}
\end{figure}
Since AlexNet contains some of the largest FC layers in modern DNNs, we performed an experimental evaluation of our MPDCompress algorithm on the AlexNet model for the full ImageNet ILSVRC-2012 dataset of 1.28 million images in 1000 different categories. Random masks with variable sparsity were applied to AlexNet FC layers FC6, FC7, and FC8 of sizes 16384$\times$4096, 4096$\times$4096, and 4096$\times$1000, respectively. We used a mini batch size of 512 images, and an initial learning rate of 3$\times$10$^{-2}$ which was reduced by a factor of 10 after every 30 epochs. The total training time for the uncompressed model was 100 epochs and we used the AlexNet model which includes batch normalization. Due to the computational complexity of network training and compression for this model/dataset combination, we implemented the training, compression, and evaluation for this case on the Amazon Web Services (AWS) platform using a P3 instance of NVIDIA Volta100 GPUs, with results provided in Fig. \ref{Top1}. To capture the trade-offs between accuracy and compression rate for real-time applications on edge devices, we ran several experiments applying different levels of compression on the fully connected layers. Fig.~\ref{Top1}a and Fig.~\ref{Top1}b show the Top 1 and Top 5 results for AlexNet for sparsity of 6.25\%, 12.5\%, and 25\% respectively. In order to maintain accuracy within 0.5\% of the non-compressed accuracy value, the compressed network was trained for twice the number of epochs compared to the uncompressed network. Ultimately, the system-level application would dictate the optimum compression rate vs. accuracy trade-off. As shown in Fig.~\ref{Top1}, for 16$\times$ network compression (6.25\% sparsity) we achieved top-1 and top-5 accuracies of 52.7$\%$ and 77.1$\%$, respectively, resulting in 4.4$\%$ and 3$\%$ accuracy loss compared to corresponding un-compressed network accuracy values. We note that 16$\times$ is an aggressive compression rate and that with further training and tuning, the accuracy loss could be further reduced. For 8$\times$ network compression (12.5\% sparsity) we achieved top-1 and top-5 accuracies of 56.4$\%$ and 79.6$\%$, respectively, resulting in 0.7$\%$ and 0.5$\%$ accuracy loss compared to corresponding un-compressed network accuracy values. We note that in  \cite{Han1}, 10$\times$ network compression was achieved without accuracy loss, but with longer training times and without potential inference speedups.  Finally, as shown in Fig. \ref{Top1}, for 4$\times$ network compression (25\% sparsity) we achieved top-1 and top-5 accuracies of 56.8$\%$ and 79.8$\%$, respectively, resulting in 0.3$\%$ accuracy loss compared to corresponding un-compressed network accuracy values.

\subsection{Inference Speedups on Hardware Platforms using MPDCompress}
We observe that our MPDCompress algorithm allows for hardware-software co-optimization at different levels. It provides a top-down approach on system design leveraging the potentials for the underlying deployment platforms such as CPUs, GPUs, FPGAs and custom ASICs as well. The inference phase is initiated with pre-optimization of the compressed network according to the deployment hardware. Structuring offers an efficient use of the computing platforms by providing techniques to allow advance scheduling and memory organization of the data. In this aspect, the time penalty for memory accesses is highly reduced for general purpose CPUs and GPUs. Further advantage is available for applications specific accelerators where the blocked structure paves the way for added parallelism and optimized architectures as well.
Consequently, aside from the compression rates and accuracy preservation, the proposed technique offers inference-mode speedup across various hardware platforms. We observe that the structured pruning of the network allows for over 4$\times$ faster operation on several GPUs with different hardware organization ~\cite{structuredDeep}. 
%
%



\begin{table}[!h]
\label{table1}
\begin{center}
\begin{tabular}{| c | c | c | c | c |}
\hline
DNN Model & \multicolumn{2}{ c |}{Evaluation Accuracy ($\%$)} & \multicolumn{2}{ c |}{Number of Parameters in FC}  \\ 
\cline{2-3}
\cline{4-5}
& MPDCompress & Non-compressed & MPDCompress & Non-compressed\\
\hline
LeNet 300-100 & 97.3 &  98.16 & 27.2k & 272k\\ \hline
Deep MNIST &99.3  & 99.3 & 322k & 3.22M\\ \hline
CIFAR10 & 85.2& 86 & 95.84k & 958.4k  \\ \hline
AlexNet (Top1) &56.4 &57.1 &11M & 87.98M\\ \hline
AlexNet (Top5) & 79.6  & 80.1 & 11M & 87.98M  \\ \hline

\end{tabular}
\end{center}
\caption{Experimental results summary achieved using the proposed MPDCompress algorithm and comparison with non-compressed results.}
\label{comparison}
\end{table}

\section{Conclusion}
%
%
DNNs have emerged as the state-of-the-art technique for machine learning tasks in computer vision applications, yet post an implementation challenge for real-time edge device applications due to their size and complexity. Previous network compression techniques based on pruning and quantization have been proposed and shown to be efficacious in reducing network size, yet can result in irregular matrix structures that are mismatched with hardware-accelerated compute platforms designed to perform the DNN matrix multiplications in a block-based way. To overcome this challenge, we have proposed MPDCompress, a DNN compression algorithm based on matrix permutation decomposition. Through in-training application of random binary masks, we achieved molding of the synaptic weight connection matrix to a sub-graph separation format, allowing for a more efficient block-based implementation and compression of the network.  We have experimentally verified our approach on several network models and datasets, including LeNet 300-100 (MNIST dataset), Deep MNIST, CIFAR10, and AlexNet (ImageNet dataset). The summaries of evaluation accuracy as well as the number of parameters with network compression using our MPDCompress algorithm and without network compression are provided in Table~\ref{comparison}. We also observe that the algorithm can offer inference speedups across various hardware platforms, with 4$\times$ faster operation on several mobile GPUs.


%



\bibliographystyle{plain}
\bibliography{egbib}

\end{document}